\documentclass{ecai} 
\usepackage{latexsym}
\usepackage{amssymb}
\usepackage{amsmath}
\usepackage{amsthm}
\usepackage{booktabs}
\usepackage{enumitem}
\usepackage{graphicx}
\usepackage{color}
\usepackage{pifont}
\usepackage{algorithmic}
\usepackage{multirow}
\usepackage[linesnumbered,ruled,vlined]{algorithm2e}

\newcommand{\BibTeX}{B\kern-.05em{\sc i\kern-.025em b}\kern-.08em\TeX}

\begin{document}

%%%%%%%%%%%%%%%%%%%%%%%%%%%%%%%%%%%%%%%%%%%%%%%%%%%%%%%%%%%%%%%%%%%%%%%%

\begin{frontmatter}

%%% Use this command to specify your submission number.
%%% In doubleblind mode, it will be printed on the first page.

\paperid{123} 

%%% Use this command to specify the title of your paper.

\title{\emph{HRS}: Hybrid Representation Framework with Scheduling Awareness for Time Series Forecasting in Crowdsourced Cloud-Edge Platforms}

%%% Use this combinations of commands to specify all authors of your 
%%% paper. Use \fnms{} and \snm{} to indicate everyone's first names 
%%% and surname. This will help the publisher with indexing the 
%%% proceedings. Please use a reasonable approximation in case your 
%%% name does not neatly split into "first names" and "surname".
%%% Specifying your ORCID digital identifier is optional. 
%%% Use the \thanks{} command to indicate one or more corresponding 
%%% authors and their email address(es). If so desired, you can specify
%%% author contributions using the \footnote{} command.

\author[A]{\fnms{Tiancheng}~\snm{Zhang}}
\author[B]{\fnms{Cheng}~\snm{Zhang}}
\author[A]{\fnms{Shuren}~\snm{Liu}}
\author[A]{\fnms{Xiaofei}~\snm{Wang}\thanks{Corresponding Author. Email: xiaofeiwang@tju.edu.cn.}}
\author[A]{\fnms{Shaoyuan}~\snm{Huang}}
\author[C]{\fnms{Wenyu}~\snm{Wang}}

\address[A]{College of Intelligence and Computing, Tianjin University}
\address[B]{Faculty of Digital Economics and Managements, Tianjin University of Finance and Economics}
\address[C]{PPIO Cloud (Shanghai) Co., Ltd.}

%%% Use this environment to include an abstract of your paper.

\begin{abstract}
With the rapid proliferation of streaming services, network load exhibits highly time-varying and bursty behavior, posing serious challenges for maintaining Quality of Service (QoS) in Crowdsourced Cloud-Edge Platforms (CCPs). While CCPs leverage  Predict-then-Schedule architecture to improve QoS and profitability, accurate load forecasting remains challenging under traffic surges. Existing methods either minimize mean absolute error, resulting in underprovisioning and potential Service Level Agreement (SLA) violations during peak periods, or adopt conservative overprovisioning strategies, which mitigate SLA risks at the expense of increased resource expenditure. To address this dilemma, we propose \emph{HRS}, a \textbf{H}ybrid \textbf{R}epresentation framework with \textbf{S}cheduling awareness that integrates numerical and image-based representations to better capture extreme load dynamics. We further introduce a Scheduling-Aware Loss (SAL) that captures the asymmetric impact of prediction errors, guiding predictions that better support scheduling decisions. Extensive experiments on four real-world datasets demonstrate that \emph{HRS} consistently outperforms ten baselines and achieves state-of-the-art performance, reducing SLA violation rates by \textbf{63.1\%} and total profit loss by \textbf{32.3\%}. Our code is available at~\cite{hrs_code}.
\end{abstract}
\end{frontmatter}

%%%%%%%%%%%%%%%%%%%%%%%%%%%%%%%%%%%%%%%%%%%%%%%%%%%%%%%%%%%%%%%%%%%%%%%%

\section{Introduction}
Streaming media services have emerged as one of the dominant forms of online content delivery, significantly shaping internet traffic patterns and user behavior. According to Sandvine's report~\cite{sandvine2024}, streaming media now accounts for nearly 40\% of global downstream traffic, and this share continues to grow. To support the rising demand for high Quality of Service (QoS), Crowdsourced Cloud-Edge Platforms (CCPs)~\cite{holistic} have emerged by aggregating idle and heterogeneous resources from distributed edge nodes. As streaming loads become increasingly dynamic and deployment technologies mature, CCPs have evolved from traditional proximity-based scheduling to more intelligent paradigms~\cite{zhang2023practical}. In particular, the Predict-then-Schedule architecture has emerged, where forecasting key metrics such as streaming traffic and server workloads guides scheduling decisions.~\cite{aaailstm, dyneformer} This enables dynamic load balancing, improves resource utilization, and enhances overall QoS and profitability .

\begin{figure}[t]
    \centering
    \includegraphics[width=0.99\columnwidth]{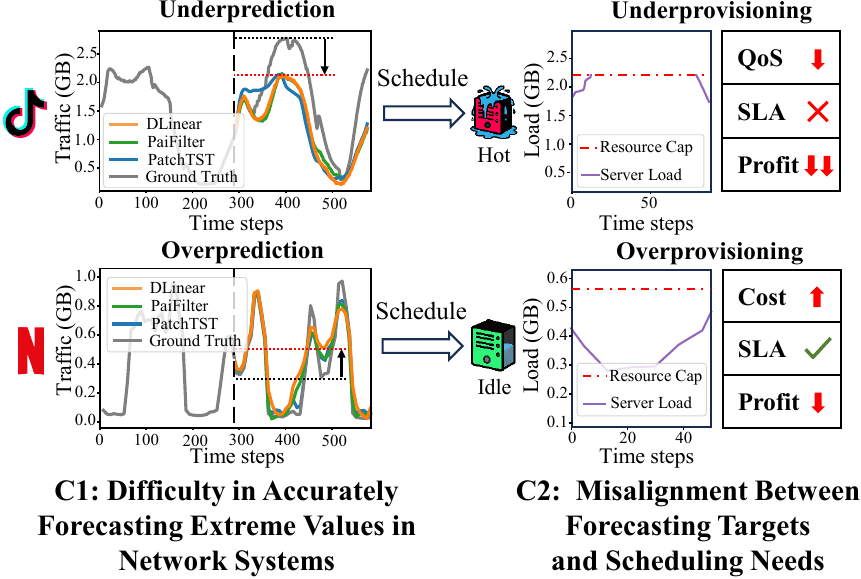}   
    \caption{Two key challenges faced by existing forecasting models in Crowdsourced Cloud-Edge Platforms.}
    \label{fig:challenge}
\end{figure}

In recent years, time series forecasting has witnessed rapid advances, with a variety of advanced models being proposed~\cite{pyraformer, informer, fedformer}. Motivated by these developments, researchers have increasingly explored integrating these models into Predict-then-Schedule frameworks~\cite{seer, igniter, oubo, sentinel}. However, despite their accuracy, the practical deployment of these models in real-world CCPs remains challenging. This paper aims to investigate the underlying factors contributing to this gap and proposes directions to bridge it.

Based on real-world observations from our collaborative CCP platform, we identify two major challenges that hinder the effective deployment of forecasting-driven scheduling: 

\textbf{Challenge 1: Difficulty in Accurately Forecasting Extreme Values in Network Systems: }In network systems, traffic bursts and short-term peaks are common due to dynamic user behaviors. Accurately forecasting peak traffic and maximum server workloads is critical for proactive scheduling and avoiding resource underprovisioning. However, existing forecasting models, which primarily rely on numerical representation, often struggle to capture extreme fluctuations. We evaluate three representative models, the linear-based DLinear~\cite{DLinear}, the frequency-based PaiFilter~\cite{FilterNet}, and the Transformer-based PatchTST~\cite{PatchTST}, on real-world streaming traffic data. As shown in Figure~\ref{fig:challenge}, although all models effectively capture the overall rising and falling trends, they consistently misestimate extreme regions, either underpredicting or overpredicting them. Such inaccuracies may lead to suboptimal scheduling decisions, especially in resource-constrained scenarios such as CCPs.

\textbf{Challenge 2: Misalignment Between Forecasting Targets and Scheduling Needs: }While minimizing absolute prediction error is a common objective in forecasting models, the primary goal in CCPs is to ensure high service quality and stable user experience. Maintaining sufficient resource provisioning to meet traffic demands is critical to avoid QoS degradation and Service-Level Agreement (SLA) violations. However, loss functions, like Mean Squared Error (MSE), treat overprediction and underprediction symmetrically, ignoring the asymmetric costs in real-world applications. As shown in Figure~\ref{fig:challenge}, underprediction can push scheduled loads beyond the resource cap, leading to degraded QoS, SLA violations, and severe penalties that significantly reduce profit. In contrast, overprediction results in resource overprovisioning and higher operational costs, while maintaining SLA and high QoS. Under highly dynamic network conditions, achieving perfect prediction is extremely challenging. Therefore, we argue that forecasting objectives should prioritize maintaining service quality and minimizing penalties, even at the cost of slight resource redundancy. 

To address these challenges, we propose \emph{HRS}, a \textbf{H}ybrid \textbf{R}epresentation time series forecasting framework with \textbf{S}cheduling awareness, designed to enhance decision-making in CCPs. \emph{HRS} introduces two key innovations:
(1) It integrates numerical and image-based representations to capture complementary shape features, improving the model’s ability to forecast extreme values. (2) It incorporates a Scheduling-Aware Loss (SAL) that accounts for the asymmetric impact of underprediction and overprediction on downstream scheduling. By selectively encouraging slight overpredictions, SAL reduces SLA violations and resource waste, ultimately improving system performance. Our main contributions are as follows:

1. We proposes a hybrid representation forecasting model with visual and numerical feature extraction branches, enabling \emph{HRS} to capture complementary shape features from time series data, and improving the accuracy of extreme value predictions.

2. We introduces a new objective through the Scheduling-Aware Loss in \emph{HRS}, which explicitly distinguishes the asymmetric impact of underprediction and overprediction on subsequent scheduling. By guiding forecasting behavior toward minimizing SLA violations and resource waste, SAL reduces the need for manual adjustments and enhances overall system performance and profitability.

3. Extensive experiments on traffic and workload forecasting across four real-world datasets demonstrate that \emph{HRS} consistently outperforms ten baselines, reducing SLA violations by \textbf{63.1\%} and improving platform profitability by decreasing profit loss by \textbf{32.3\%}.

\section{Related Works}\label{rela}
\textbf{Evolution of Time Series Forecasting Models: }Traditional models, such as LSTM~\cite{lstm}, have been widely adopted for their ability to capture temporal patterns. With the introduction of Transformer architectures, a new generation of forecasting models emerged, including Informer~\cite{informer}, Autoformer~\cite{autoformer}, and Pyraformer~\cite{pyraformer}, which achieved impressive results across diverse scenarios. More recently, lightweight linear models such as DLinear~\cite{DLinear} and state-space models like Mamba~\cite{mamba} have gained attention, offering competitive performance with lower computational complexity. However, despite their success, these models often struggle in highly dynamic scenarios, particularly in forecasting short-term fluctuations and extreme values, factors that are critical in network systems where traffic spikes can significantly impact scheduling and service delivery.

\textbf{Representation Learning for Time Series Forecasting: }Most existing forecasting models operate purely within the numerical domain, learning temporal dependencies from raw vector. Recent advances have proposed alternative perspectives, such as frequency-domain transformations~\cite{fedformer, FilterNet} and patch-based segmentation techniques~\cite{PatchTST, ts}, to better exploit temporal information. Nevertheless, these approaches remain confined to numerical representation and may struggle to capture localized shape patterns, such as sharp peaks or abrupt drops. Parallel efforts have explored transforming time series data into visual representations, such as visualTrans~\cite{visualTrans}, mainly for classification tasks rather than for regression or forecasting. 

\textbf{Scheduling-Aware Forecasting Objectives: }
Most existing forecasting models optimize symmetric objectives like MSE, treating overprediction and underprediction equally. In practice, however, prediction errors in CCPs have asymmetric consequences, with each direction impacting system performance differently. While studies like~\cite{deepcog} and~\cite{deepcogac}, have begun integrating forecasting models with scheduling needs, their approaches typically rely on simple penalty biases or coarse adjustment mechanisms. These methods lack fine-grained modeling of the relationship between forecasted values and associated operational costs.

\begin{figure*}[t]
    \centering
    \includegraphics[width=0.9\linewidth]{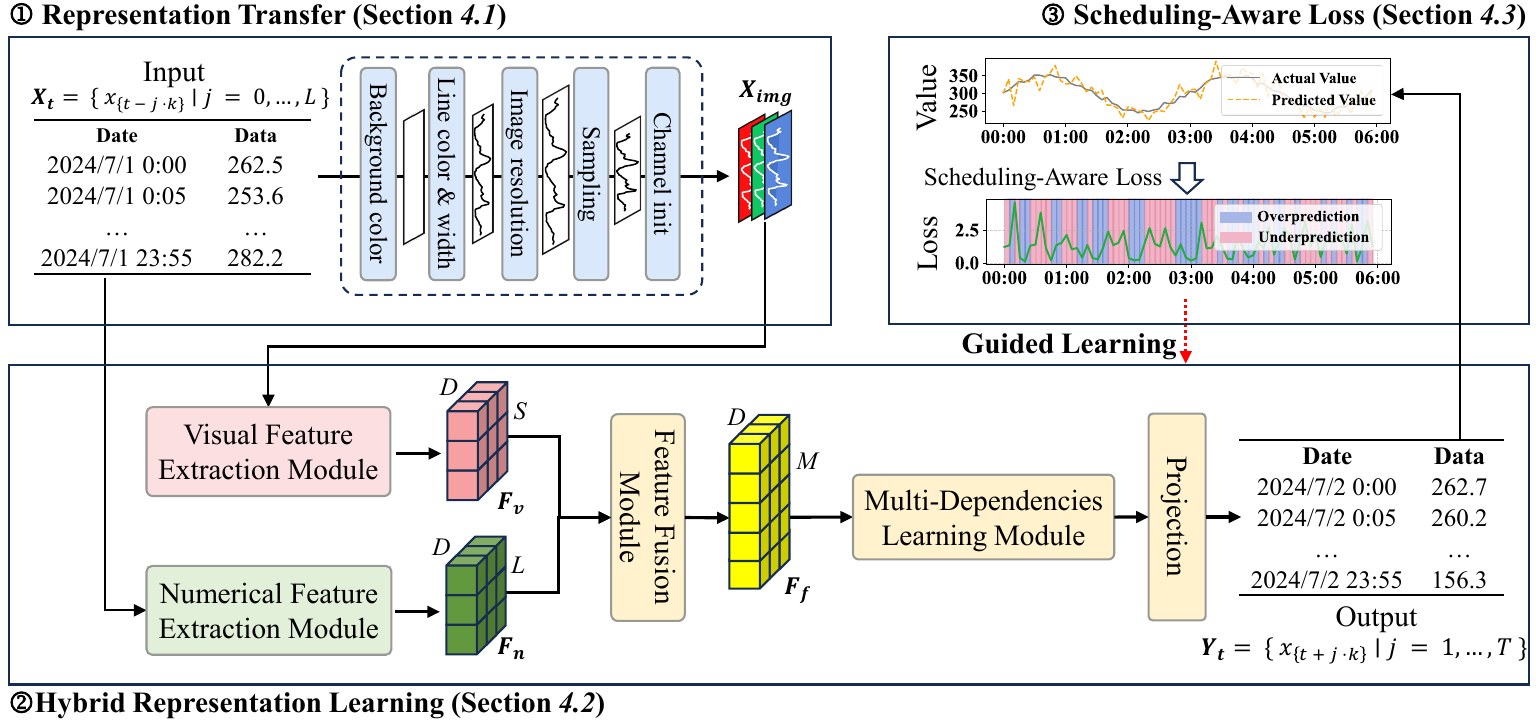}   
    \caption{Overview of the \emph{HRS}. \ding{172} Numerical time series are transformed into image-based representations. \ding{173} Hybrid Representation Learning process extracts features from both numerical and image inputs and fuses them for future forecasting. \ding{174} Scheduling-Aware Loss guides learning by penalizing over- and under-prediction asymmetrically.}
    \label{fig:overview}
\end{figure*}

\section{Preliminary}
To achieve effective scheduling, CCP relies primarily on two key indicators, streaming traffic and server load, to ensure that services are allocated to appropriate resources, thus enhancing both QoS and overall profitability. These indicators are typically collected by a sampler at fixed time intervals $k$. The forecasting task aims to predict the next $T$ intervals based on the previous $L$ historical observations. Formally, at the time step $t$, the input of the model is defined as $X_t = \{ x_{t - j \cdot k} \mid j = 0, \dots, L \}$, and the goal is to forecast the future sequence $Y_t = \{ x_{t + j \cdot k} \mid j = 1, \dots, T \}$.

\section{Methodology}
In this section, we describe the proposed \emph{HRS} model and how it is designed to address the two key challenges introduced in Section 1. We first provide an overview of the model, followed by detailed explanations of its core modules and training objectives.

\emph{HRS} consists of three main components, as illustrated in Figure~\ref{fig:overview}. 
\ding{172} The input time series $X_t$ is transformed into images. This image-based representation enhances the model’s ability to capture localized fluctuations and sharp peaks in the data. \ding{173} These image-based representations, together with the original numerical inputs, are fed into the Hybrid Representation Learning component. This component extracts features from different representations through separate branches, which are then fused into a unified hybrid representation $F_{f}$. A Multi-Dependencies MLP module further captures temporal dependencies from $F_{f}$ and generates the prediction $Y_t$, effectively capturing global trends while remaining sensitive to fluctuations and extreme values.
\ding{174} Finally, the model is trained with a Scheduling-Aware Loss Function that explicitly distinguishes the asymmetric impact of prediction errors. This guides the learning process to produce forecasts that better align with real-world scheduling objectives.

\begin{figure}[ht]
    \centering
    \includegraphics[width=0.95\columnwidth]{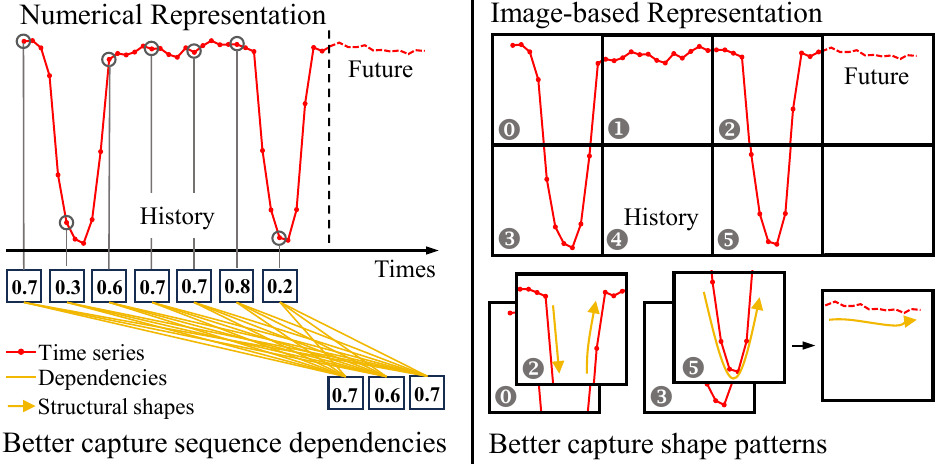}
    \caption{Complementary strengths of numerical and image-based representation 
 in time series forecasting tasks.}
    \label{fig:multimodal}
\end{figure}

\subsection{Representation Transfer}\label{sec:image}

Extreme regions in network datasets often exhibit sharp peaks, deep valleys, or sudden transitions. These patterns are difficult to precisely characterize using numerical data alone, as it primarily reflects sequence dependencies. To better capture internal associations and extreme regions in time series, we transform the input from the commonly used numerical representation into an image-based representation. This image-based representation is designed to complement the numerical representation, as illustrated in Figure~\ref{fig:multimodal}.

 To improve efficiency, the image conversion is implemented during the data loading phase. To convert time series into meaningful image representations, certain visual rendering properties, such as background color, line width, and resolution, need to be defined. As shown in Algorithm 1 in Appendix~\ref{algo}, each feature in the time series is transformed into a 3-channel image using OpenCV line drawing. The series is first normalized via min-max scaling and then rendered as a polyline on an RGB canvas with a fixed background. The resulting image is stored in the output tensor $X_{\text{img}} \in \mathbb{R}^{3M \times H \times W}$.

\subsection{Hybrid Representation Learning}
In this section, we describe the main modules of our proposed Hybrid Representation Learning component.

\begin{figure}[hbtp]
    \centering
    \includegraphics[width=0.9\columnwidth]{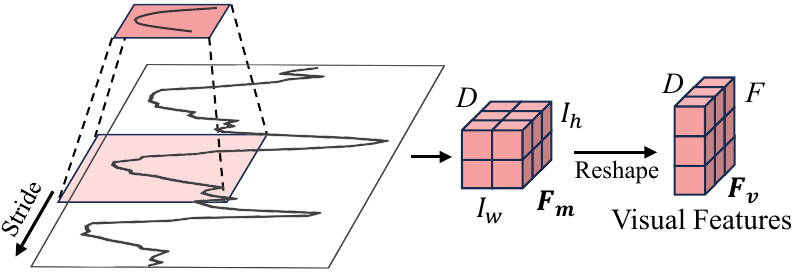}   
    \caption{VFEM applies 2D convolution to image-based representation to extract local shape patterns.}
    \label{fig:VFEL}
\end{figure}

\textbf{Visual Feature Extraction Module (VFEM).} VFEM is designed to capture fine-grained shape patterns from image-based representations of time series data. As shown in Figure~\ref{fig:VFEL}, it applies 2D convolutional layers with kernel size $(k_h, k_w)$ and stride $(s_h, s_w)$ to captures salient shape structures while compressing redundant spatial information. Similar to the Patch~\cite{vittrans} technique, this design allows \emph{HRS} to preserve critical local temporal semantics and support longer input sequences without causing tensor dimensional explosion. 

Formally, VFEM transforms the image input $X_{img}$ into a shape feature map $S_m$, which is then reshaped for integration with features extracted from the numerical representation:
\begin{align}
    S_m & = \operatorname{Conv2d}\left(X_{img}\right), \\
    F_v & = \operatorname{Reshape}\left(S_m\right).
\end{align}
Here, $S_m \in \mathbb{R}^{I_w \times I_h \times D}$, where $I_w, I_h$, and $D$ represent the width, height, and dimensionality of the feature map after convolution. $S_m$ is reshaped into $F_v \in \mathbb{R}^{V \times D}$, where $V = I_w \times I_h$.

\begin{figure}[htbp]
    \centering
    \includegraphics[width=0.9\columnwidth]{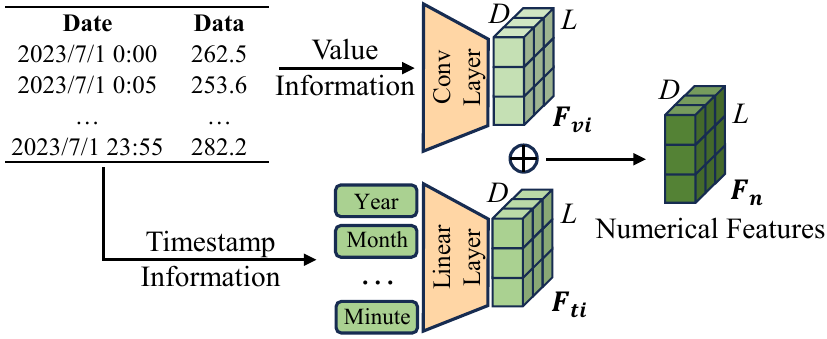}   
    \caption{NFEM: Sequential values and timestamps are processed in parallel and fused into numerical features.}
    \label{fig:NFEL}
\end{figure}

\textbf{Numerical Feature Extraction Module (NFEM).}  
NFEM captures temporal dependencies from numerical time series through two key branches, as illustrated in Figure~\ref{fig:NFEL}. Given an input numerical sequence with timestamps and values, denoted as $X_{\text{num}}[\text{Date}] \in \mathbb{R}^{L}$ and $X_{\text{num}}[\text{Data}] \in \mathbb{R}^{L}$, the first branch applies a 1D convolution to transform $X_{\text{num}}[\text{Data}]$ into $T_{vl} \in \mathbb{R}^{L \times D}$, where $L$ is the sequence length and $D$ the embedding dimension, consistent with VFEM. This transformation enables the model to capture complex temporal dependencies in a higher-dimensional feature space.

Many real-world data exhibit multi-periodic patterns driven by social behaviors across different time scales (e.g., monthly, weekly, daily)~\cite{lspattern}. For instance, streaming traffic often peaks during holidays (monthly), weekends (weekly), and nighttime (daily). To explicitly model such periodicity, the second branch extracts structured temporal features from timestamps. The timestamp $X_{\text{num}}[\text{Date}]$ is decomposed into multiple time fields (e.g., year, month) and mapped through a linear transformation to produce $F_{ti} \in \mathbb{R}^{L \times D}$.

The final output $F_n$ is obtained by integrating temporal dependencies and structured periodic signals through element-wise addition:
\begin{align}
    F_{vi} & = \operatorname{Conv1d}\left(X_\text{num}[\text{Data}]\right), \\
    F_{ti} & = \operatorname{Linear}\left(\operatorname{Decompose}(X_\text{num}[\text{Date}])\right), \\
    F_n & = F_{vi} + F_{ti}.
\end{align}

\textbf{Feature Fusion Module (FFM).}  
A key challenge in hybrid representation learning is effectively integrating complementary information from different representations. As previously discussed, numerical features primarily captures periodicity and long-term sequential dependencies, whereas the visual features emphasizes localized shapes. To fuse these strengths, we introduce FFM, which concatenates the visual and numerical features along the feature dimension and projects them into an $M$-dimensional space via a linear layer. This operation enables the model to adaptively learn cross-representation interactions. The fusion process is defined as:

\begin{equation}
    F_f = W_{ff} \left[\; F_v \; ; \; F_n \; \right] + b, \quad W_{ff} \in \mathbb{R}^{M \times (V+L)},
\end{equation}
where $\left[\; F_v \; ; \; F_n \; \right]$ denotes the concatenation of $F_v$ and $F_n$ along the feature dimension, $W_{ff}$ is the learnable weight matrix, and $b$ is the bias term. Here, $V$ and $L$ represent the dimensions of the visual and numerical feature vectors, respectively.

\begin{figure}[htbp]
    \centering
    \includegraphics[width=0.9\columnwidth]{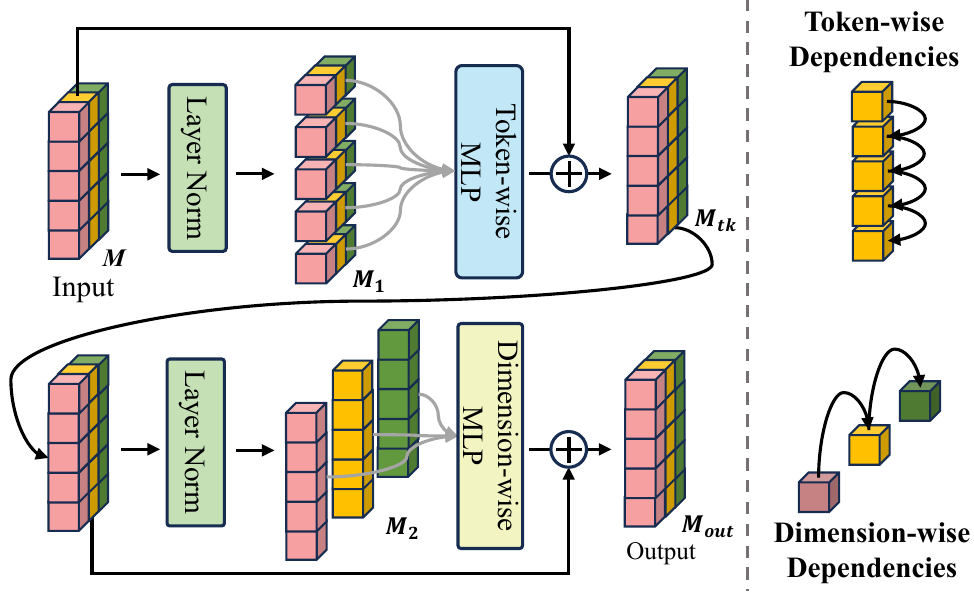}   
    \caption{MDM: Learning multi-dependencies in hybrid representation.}
    \label{fig:MultiMLP}
\end{figure}

\textbf{Multi-Dependency Learning Module (MDM).} MDM is designed to capture complex interactions within the fused feature of different representations by applying a combination of token-wise and dimension-wise modeling. As illustrated in Figure~\ref{fig:MultiMLP}, the fused feature $F_f$ first undergoes layer normalization and is then processed vertically through a token-wise MLP, which models dependencies across tokens along the sequence dimension.  

The output from the token-wise MLP is combined with the original input via a residual connection, then normalized and passed horizontally into a dimension-wise MLP. The dimension-wise MLP captures dependencies across feature dimensions within each token. Given that the input is a fused feature of visual and numerical representations, MDM is designed to fully model cross-representation dependencies. Formally, the computation flow of MDM is as follows:
\begin{align}
    M_1 & = \operatorname{LayerNorm}\left(F_f\right), \\
    M_{tk} & = \operatorname{Token-wise} \operatorname{MLP}(M_1), \\
    M_2 & = \operatorname{LayerNorm}(M_{tk} + F_f), \\
    M_{out} & = \operatorname{Dimension-wise} \operatorname{MLP}(M_2),
\end{align}
where $F_f$ denotes the fused feature, $M_{tk}$ represents the intermediate output from the token-wise MLP, and $M_{out}$ is the final output after dimension-wise modeling. The residual connections and normalization layers are introduced to stabilize training.

\subsection{Scheduling-Aware Loss}

\begin{figure*}[htbp]
    \centering
    \includegraphics[width=0.8\linewidth]{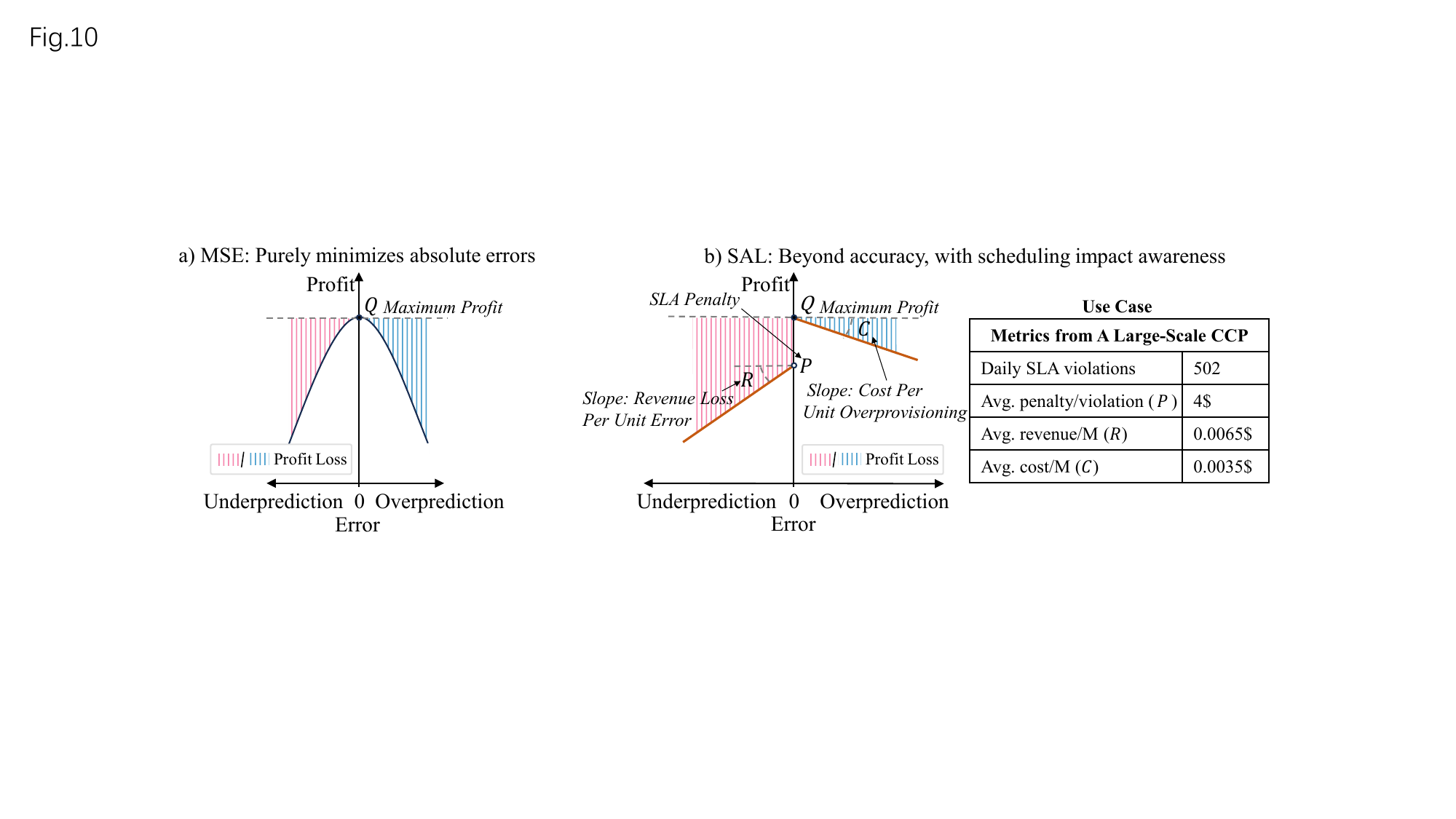}   
    \caption{Comparison Between MSE and SAL. Unlike MSE, which treats all errors equally, SAL models the asymmetric impact of prediction errors on scheduling, guiding safer and more scheduling-aware predictions.}
    \label{fig:loss}
\end{figure*}

As previously discussed, conventional loss functions such as MSE and MAE treat all deviations equally, failing to reflect the asymmetric impact of prediction errors on downstream scheduling decisions in CCPs. In practice, underprediction may cause overloads, QoS drops and SLA violations, while overprediction typically ensures QoS and avoids SLA violations, but leads to increased resource costs.

To address this, we introduce a Scheduling-Aware Loss (SAL), which uses profit impact as a unified metric to model the consequences of overprediction and underprediction. SAL encourages forecasting behavior that minimizes SLA violations and resource waste—two key factors that directly affect downstream scheduling objectives, including service quality and platform profitability. The formulation of SAL is as follows:
\begin{equation}
\textbf{SAL}(Y, \hat{Y}) =
\begin{cases}
R \cdot (Y - \hat{Y}) + P , & \text{if } \hat{Y} < Y. \\
0, & \text{if }  \hat{Y} = Y. \\
C \cdot (\hat{Y} - Y), & \text{if } \hat{Y} > Y.
\end{cases}
\end{equation}

Where $\hat{Y}$ and ${Y}$ denote the predicted and actual values. $R$ is the revenue coefficient, indicating the profit CCPs earn for servicing one unit of traffic. $P$ is the SLA penalty when underprediction leads to SLA violations. $C$ is the per-unit resource cost due to overprediction.

The scheduling-optimal baseline assumes a perfect forecast, where supply precisely meets demand and yields the theoretical maximum profit $Q$, as illustrated in Figure~\ref{fig:loss}. However, when underprediction occurs (e.g., $\hat{Y} < Y$ for traffic), the system may assign tasks to already high-load servers or under-allocate bandwidth, causing SLA violations and lost service opportunities—incurring both $P$ and $R \cdot (Y - \hat{Y})$ in loss. On the other hand, overprediction may lead to overprovisioning, resulting in unnecessary resource costs, incurring a loss of $C \cdot (\hat{Y} - Y)$. A similar effect occurs for server workload prediction, where underprediction may overload servers, and overprediction may lead to underutilization and resource inefficiency.

Importantly, as illustrated in Figure~\ref{fig:loss}.b), real-world CCP metrics (detailed in ~\ref{mfac}) reveals that the cost of overprovisioning ($C$) is significantly lower than the expected revenue per unit traffic ($R$), and much lower than SLA penalties ($P$), satisfying $C < R \ll P$. This asymmetry motivates SAL to favor safe overprovisioning when perfect accuracy is unattainable, especially under dynamic network conditions, to reduce SLA violations and mitigate profit loss. 

\section{Experiment}
In this section, we present comprehensive experimental results to evaluate the effectiveness of the proposed \emph{HRS}. We first report its performance on real-world data, followed by ablation studies to assess the contributions of each component. We then analyze the computational complexity of \emph{HRS} to evaluate its scalability and deployment feasibility. Finally, we demonstrate its practical value through a real-world use case in~\ref{usecase} and visualization analyses in~\ref{visu}, highlighting its advantage in forecasting extreme values. We also evaluate its generalization to related domains in Appendix~\ref{gen}.

\begin{table}[htbp]
  \centering
  \caption{Datasets Details}
    \begin{tabular}{ccccc}
    \hline
    Datasets ($\mathcal{D}$) & Types & Timesteps & Frequencies \\
    \hline
    ECT1    & Streaming Traffic & 74304  & 5 mins \\
    ECT2    & Streaming Traffic & 74304  & 5 mins \\
    ECW08    & Server Workload & 797×720  & 1 hour \\
    ECW09    & Server Workload & 1022×720  & 1 hour \\
    
    \hline
    \end{tabular}%
  \label{tab:datasets}%
\end{table}%

\textbf{Datasets. }To evaluate \textit{HRS} in real-world settings, we collected data from a commercial CCP with over 5,000 distributed servers more than 40 applications. We anonymized traffic data from two of its most popular services, covering the period from July 1, 2023, to March 14, 2024, resulting in the \textbf{E}dge \textbf{C}loud \textbf{T}raffic datasets, ECT1 and ECT2. Additionally, we used the public server workload datasets ECW08 and ECW09 from DynEformer~\cite{dyneformer}, which record 720 hours of upload bandwidth from over 700 edge servers (Aug–Sep 2022). Dataset details are shown in Table~\ref{tab:datasets}.

\begin{table*}[tbp]
  \centering
  \setlength{\tabcolsep}{4 pt}
  \caption{Model performance on the ECT and ECW datasets. \textbf{Bold} indicates the best, and \underline{underlined} the second-best.}
    \begin{tabular}{c|c|c|ccc|ccc|ccc|ccc}
    \toprule
    \multirow{2}[4]{*}{\rotatebox{90}{Types}} & \multirow{2}[4]{*}{Models} & $\mathcal{D}$ →   & \multicolumn{3}{c|}{ECT1 (5 mins)} & \multicolumn{3}{c|}{ECT2 (5 mins)} & \multicolumn{3}{c|}{ECW08 (1 hour)} & \multicolumn{3}{c}{ECW09 (1 hour)} \\
\cmidrule{3-15}          &       & $T$ →   & 288   & 576   & 864   & 288   & 576   & 864   & 24    & 48    & 72    & 24    & 48    & 72 \\
    \midrule
    \multicolumn{2}{c|}{\emph{HRS} (Ours)} & APL   & \textbf{1.555 } & \textbf{1.761 } & \textbf{1.703 } & \textbf{0.974 } & \textbf{1.274 } & \textbf{1.209 } & \textbf{1.393 } & \textbf{1.672 } & \textbf{1.535 } & \textbf{0.940 } & \textbf{1.085 } & \textbf{1.228 } \\
    \midrule
    \multirow{10}[20]{*}{\rotatebox{90}{Trained with MSE}} & Timemachine (ECAI 24) & APL   & 3.073  & 3.128  & 3.356  & 2.445  & 2.641  & 2.457  & 2.767  & 2.767  & 2.744  & 2.476  & 2.574  & 2.694  \\ 
\cmidrule{2-15}          & TexFilter (NeurIPS 24) & APL   & 3.443  & 3.109  & 3.406  & 2.611  & 2.677  & 2.725  & 2.848  & 2.808  & 3.137  & 1.770  & 2.122  & 2.522  \\
\cmidrule{2-15}          & PaiFilter (NeurIPS 24) & APL   & 3.106  & 3.116  & 3.350  & 2.759  & 2.868  & 2.957  & 2.893  & 2.875  & 2.673  & 2.351  & 2.441  & 2.510  \\
\cmidrule{2-15}          & TimeMixer (ICLR 24) & APL   & 2.826  & 3.238  & 3.041  & 1.740  & 2.805  & 2.811  & 3.522  & 3.460  & 3.460  & 2.876  & 2.685  & 2.985  \\
\cmidrule{2-15}          & iTransformer (ICLR 24) & APL   & 2.941  & 3.397  & 3.294  & 2.750  & 2.944  & 3.065  & 3.906  & 3.283  & 3.489  & 2.523  & 3.124  & 2.724  \\
\cmidrule{2-15}          & TSMixer (TMLR 23) & APL   & 2.505  & 3.009  & 2.714  & 2.780  & 2.958  & 3.038  & 2.991  & 2.969  & 2.823  & 2.592  & 2.360  & 2.231  \\
\cmidrule{2-15}          & PatchTST (ICLR 23) & APL   & 3.512  & 3.168  & 3.612  & 2.114  & 2.064  & 2.655  & 2.746  & 2.745  & 2.561  & 2.728  & 2.325  & 2.453  \\
\cmidrule{2-15}          & Timesnet (ICLR 23) & APL   & 4.254  & 3.862  & 3.102  & 2.826  & 2.220  & 3.764  & 3.388  & 3.868  & 3.958  & 3.289  & 3.840  & 3.873  \\
\cmidrule{2-15}          & DLinear (AAAI 23) & APL   & 2.350  & 2.598  & 2.756  & 3.118  & 3.437  & 3.466  & 2.953  & 2.986  & 2.851  & 2.200  & 2.201  & 2.360  \\
\cmidrule{2-15}          & GPT4TS (NeurIPS 23) & APL   & 3.476 & 3.372 & 3.428 & 2.833 & 2.676 & 2.186 & 2.942 & 3.130  & 2.923 & 3.027 & 2.407 & 2.265 \\
    \midrule
    \multirow{9}[18]{*}{\rotatebox{90}{Trained with SAL}} & TimeMixer & APL   & 1.686  & 2.254  & 2.381  & \underline{1.197}  & 1.975  & 1.791  & 2.399  & 2.755  & 2.703  & 1.604  & 1.885  & 2.144  \\
\cmidrule{2-15}          & TexFLiter & APL   & 2.143  & 2.325  & 2.505  & 1.611  & \underline{1.601}  & \underline{1.479}  & 2.115  & 2.155  & \underline{1.679}  & 1.591  & 1.792  & 1.399  \\
\cmidrule{2-15}          & PaiFilter & APL   & 2.175  & 2.230  & 2.224  & 1.931  & 2.325  & 2.289  & 2.144  & 2.167  & 2.254  & 1.450  & 1.612  & 1.306  \\
\cmidrule{2-15}          & Timemachine & APL   & 2.090  & 2.247  & 2.632  & 1.949  & 2.042  & 2.157  & 2.137  & \underline{2.093}  & 1.985  & 1.336  & 1.279  & 1.388  \\
\cmidrule{2-15}          & iTransformer & APL   & 1.957  & 2.334  & 2.608  & 2.286  & 2.499  & 2.644  & 2.245  & 2.327  & 2.335  & 1.230  & 1.544  & 1.442  \\
\cmidrule{2-15}          & TSMixer & APL   & 1.711  & \underline{1.843}  & 2.033  & 2.069  & 2.327  & 2.466  & \underline{1.962}  & 2.326  & 2.336  & 1.438  & 1.272  & 1.322  \\
\cmidrule{2-15}          & PatchTST & APL   & 2.462  & 2.386  & 2.559  & 1.515  & 1.842  & 1.732  & 2.040  & 2.108  & 2.089  & \underline{0.983}  & 1.627  & \underline{1.295}  \\
\cmidrule{2-15}          & Timesnet & APL   & 3.784  & 3.272  & 4.268  & 2.617  & 3.324  & 3.531  & 2.653  & 2.699  & 2.476  & 2.838  & 1.861  & 3.967  \\
\cmidrule{2-15}          & DLinear & APL   & \underline{1.680}  & 1.948  & 2.088  & 2.099  & 2.320  & 2.402  & 2.427  & 2.426  & 2.205  & 1.621  & 1.690  & 1.794  \\
\cmidrule{2-15}          & GPT4TS & APL   & 2.289 & 2.332 & \underline{1.828} & 2.156 & 1.889 & 2.266 & 2.333 & 2.427 & 2.390  & 1.048 & \underline{1.236} & 1.419 \\
    \bottomrule
    \end{tabular}%
  \label{tab:result}%
\end{table*}%

\textbf{Baseline. }To evaluate the performance of \emph{HRS}, we compare it against ten representative baselines. These include frequency-based models such as PaiFilter and TexFilter~\cite{FilterNet}, linear models including TimeMixer~\cite{timemixer}, DLinear~\cite{DLinear}, and TSMixer~\cite{ts}, as well as transformer-based models like iTransformer~\cite{itransformer} and PatchTST~\cite{PatchTST}. We also include TimeMachine~\cite{timemachine}, a state-space model based on the Mamba, TimesNet~\cite{timesnet}, a CNN-based model and GPT4TS~\cite{GPT4TS}, a forecasting model built on pretrained large language models. These models have demonstrated strong generalization ability and competitive performance across various time series forecasting tasks. 

\textbf{Implementation Details. }To meet the needs of different applications, we evaluate predictions for various lengths: 1-day, 2-day, and 3-day forecasts. For ECT (5-minute interval), prediction lengths are $T = \{288, 576, 864\}$; for ECW (1-hour interval), $T = \{24, 48, 72\}$. Since different baselines use varying input lengths, we standardize the input length $L = P$ for fair comparison.

SAL is based on three key parameters derived from average values calculated from real CCP data after desensitization: the revenue coefficient $R = 0.0065$, the cost coefficient $C = 0.0035$, and the penalty $P = 4$. The ECT dataset is split into 7:1:2 for training, validation, and test sets, while the ECW dataset follows a 6:2:2 split, as in the original work~\cite{dyneformer}. All experiments use a batch size of 32. The parameters for each baseline follow the settings from the original work for similar lengths. All experiments were implemented in PyTorch \cite{pytorch} and conducted on an NVIDIA V100 GPU machine.

\subsection{Main Result}
Table~\ref{tab:result} presents the performance of \emph{HRS} on the ECT and ECW datasets. To evaluate the impact of prediction quality on downstream scheduling, we use SAL defined in Eq.~(11) as a unified profit-based metric. Specifically, we report APL (Average Profit Loss), which measures the average profit loss per test point; a lower APL indicates less profit loss occurs, meaning the prediction are better support reasonable scheduling decisions. Each baseline is evaluated under two training objectives: standard MSE (from original implementations) and SAL (proposed in this work).

Baselines in their original implementations generally result in higher profit loss, indicating that their prediction accuracy is insufficient to support reasonable scheduling decisions in CCPs. One possible reason is the tendency toward underprediction, which results in under-allocation of resources. This, in turn, degrades QoS, causes SLA violations, and incurs penalty losses, ultimately reducing overall platform performance and profitability. In contrast, when trained with SAL, baselines are guided to recognize the asymmetric impact of prediction errors, achieving an average APL improvement of 27.9\% across all datasets. By encouraging safer forecasting behavior, SAL helps reduce SLA violations and improve platform efficiency under imperfect prediction conditions.

Notably, even all baselines are trained with SAL, \emph{HRS} consistently achieves superior performance across all datasets. It yields an average APL reduction of 25.5\% and 43.9\% on ECT1 and ECT2, respectively, and 32.0\% and 27.9\% on ECW08 and ECW09. While some baselines perform competitively on individual datasets, \emph{HRS} outperforms the second-best model by an average margin of 12.9\% across all tasks. This consistent improvement stems from its hybrid representation design, which integrates numerical and image-based representation to capture complementary patterns.

\begin{figure}[htbp]
    \centering
    \includegraphics[width=0.99\columnwidth]{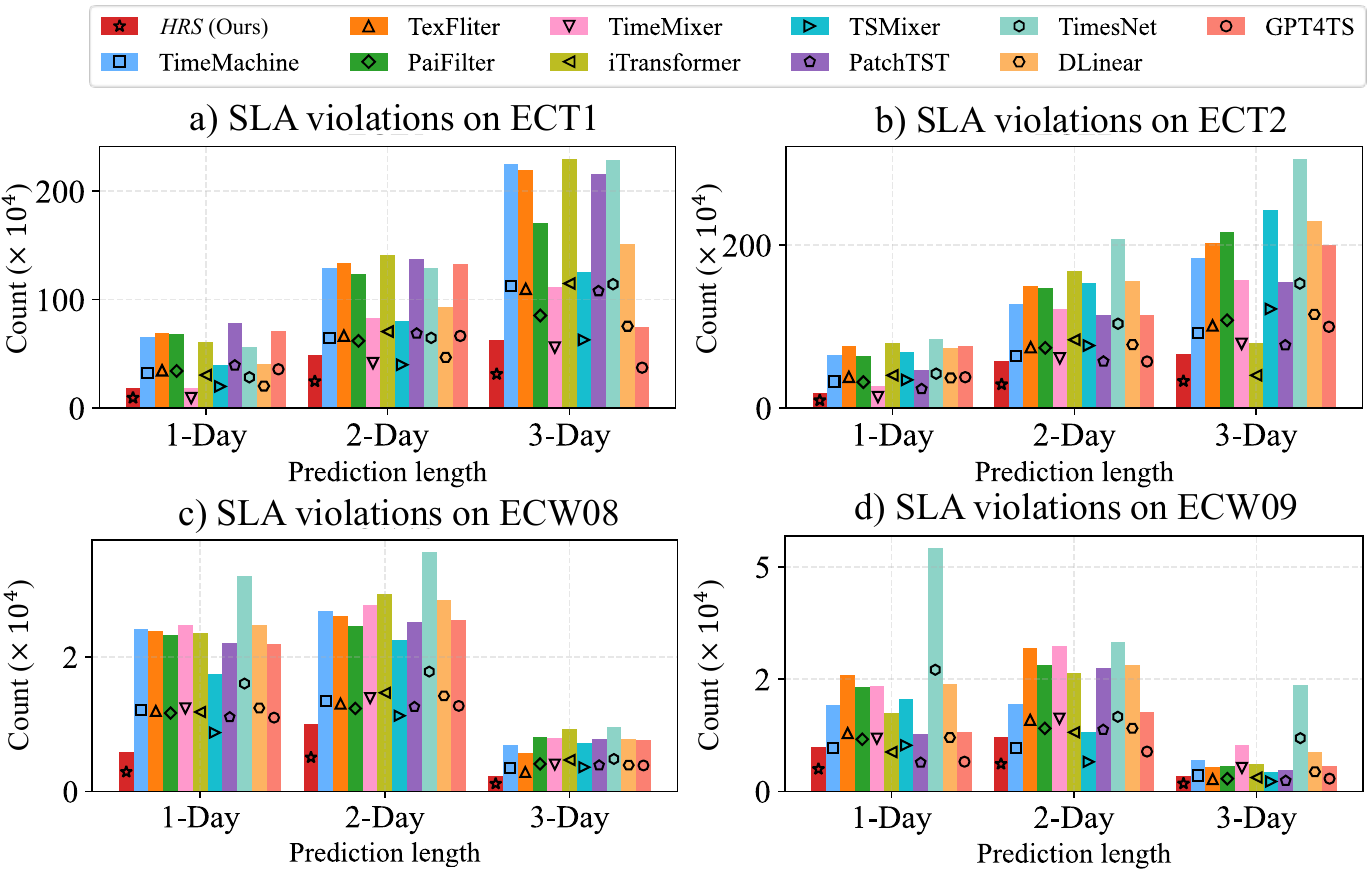}   
    \caption{SLA violations under different lengths on ECT and ECW.}
    \label{fig:SLA_vialation}
\end{figure}

\textbf{Analysis of SLA Violations. }Figure~\ref{fig:SLA_vialation} shows the SLA violations of each model on ECT and ECW, highlighting cases where predicted traffic or workload falls short of actual values, leading to insufficient resource allocation. All models are trained with SAL. Across three prediction lengths, \emph{HRS} consistently yields the fewest SLA violations in both traffic and workload forecasting tasks. It achieves an average reduction in SLA violations of 61.8\%, 66.4\%, 72.9\%, and 51.2\% on ECT1, ECT2, ECW08, and ECW09, respectively. This demonstrates \emph{HRS}’s effectiveness in providing more reliable predictions, ultimately supporting better scheduling decisions.

\begin{figure*}[htbp]
    \centering
    \includegraphics[width=0.99\linewidth]{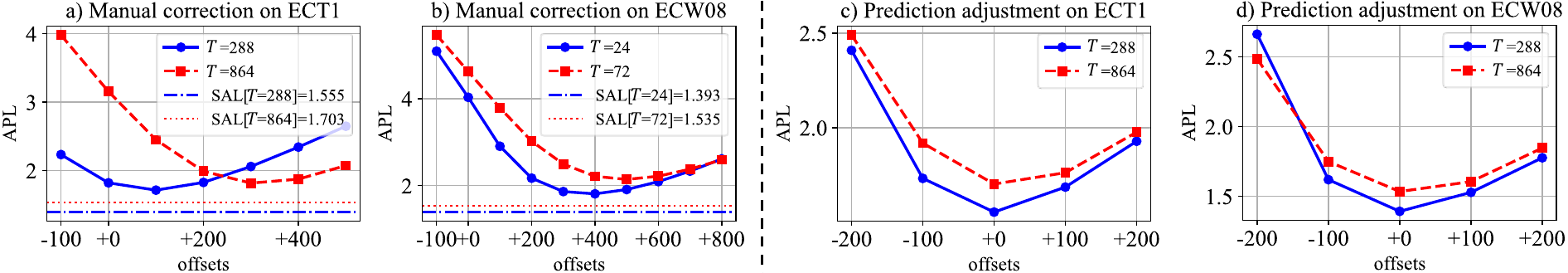}   
    \caption{a)-b) Comparison between SAL-trained \emph{HRS} and MSE-trained \emph{HRS} with manually introduced biases under different offset settings. $T$ is the prediction length, and SAL[$T$=X] refers to models trained with SAL at length X. c)–d) Evaluation of directly adjusted predictions from SAL-trained \emph{HRS}.
SAL consistently achieves lower APL, demonstrating its effectiveness in balancing overprovisioning and SLA violations.}
    \label{fig:abla_pol}
\end{figure*}

\subsection{Ablation Study and Hyperparameter Sensitivity Analysis}

\textbf{Ablation Study on Hybrid Representation Learning. }We conduct ablation experiments by systematically removing key modules from the Hybrid Representation Learning component, as shown in Table~\ref{tab:abla}. 

\textbf{w/o VFEM: }We remove the visual feature extraction branch and rely solely on numerical representation. To ensure fairness, we maintain the same feature dimensionality. Performance drops across all prediction lengths, highlighting the importance of image-based representation in capturing localized shape patterns.

\textbf{w/o NFEM: }We remove the NFEM and retain only the image-based representation. The feature dimensionality is kept consistent to rule out the effect of reduced model capacity. The results show a noticeable decline, indicating that numerical representation provides richer temporal information and is essential for accurate forecasting.

\textbf{w/o FFM: }To assess the impact of FFM, we replace it with a simple concatenation of numerical and visual features. The degraded performance suggests that naïve concatenation fails to effectively integrate complementary information, underscoring the necessity of a fusion mechanism to capture cross-representation interactions.

\textbf{w/o MDM: }The MDM module captures dependencies from both token-wise and dimension-wise perspectives using MLPs. To validate its effectiveness, we replace MDM with two sequentially connected MLP layers for comparison. The consistent performance drop confirms the effectiveness of token-wise and dimension-wise modeling in enhancing predictive accuracy.

\begin{table}[htbp]
  \centering
  \setlength{\tabcolsep}{2.5pt}
  \caption{Ablation Study on the Hybrid Representation Learning}
    \begin{tabular}{c|c|c|cccc}
    \toprule
    \multicolumn{2}{c|}{Models →} & \emph{HRS} & w/o VFEM & w/o NFEM & w/o FFM   & w/o MDM \\
    \midrule
    {$\mathcal{D}$} & {$T$} & APL & APL & APL & APL   & APL \\
    \midrule
    \multirow{2}[2]{*}{ECT1} & 288   & \textbf{1.555 } & 3.419 & 3.428 & 2.540 & 3.468 \\
          & 864   & \textbf{1.703 } & 3.281 & 3.514 & 3.782 & 4.769 \\
    \midrule
    \multirow{2}[2]{*}{ECW08} & 24    & \textbf{1.393} & 1.760 & 6.166 & 1.573 & 2.487 \\
          & 72    & \textbf{1.535} & 5.001 & 3.673 & 3.182 & 2.382 \\
    \bottomrule
    \end{tabular}%
  \label{tab:abla}%
\end{table}%

\textbf{Ablation Study on SAL. }To validate the effectiveness of SAL, we conduct two ablation experiments. First, we train \emph{HRS} using MSE loss and apply manual bias offsets to its predictions, simulating the common industry practice of post-hoc correction. As shown in Figure~\ref{fig:abla_pol}.a) and b), the profit loss across different forecasting tasks is consistently higher compared to training directly with SAL, suggesting that SAL inherently provides a better balance between overprovisioning and SLA violations. Second, we apply controlled offsets to the predictions generated by \emph{HRS} trained with SAL. As seen in Figure~\ref{fig:abla_pol}.c) and d), any deviation results in bigger profit loss, demonstrating that SAL enables the model to learn an optimal trade-off, supporting more effective scheduling decisions.

\textbf{Sensitivity of representation transfer settings.}
We assess the impact of visual rendering factors, including line width ($lw$), line color ($lc$), and background color ($bc$), where $lc$ and $bc$ vary across RGB settings: r (1,0,0), g (0,1,0), and b (0,0,1). For brevity, results in Table~\ref{tab:sen_image} are summarized using the coefficient of variation (CV). All CV values are below 0.2, indicating that \emph{HRS} is robust to these settings and consistently captures shape patterns across visual variations.

\begin{table}[htbp]
  \centering
  \caption{ Sensitivity analysis of representation transfer settings}
  \setlength{\tabcolsep}{4pt}
    \begin{tabular}{c|c|c|c|c}
    \toprule
    \multicolumn{2}{c|}{Settings →} & $lw$ = \{1, 2, 3\} & $lc$ = \{r, g, b\} & $bc$ = \{r, g, b\} \\
    \midrule
    $\mathcal{D}$     & $T$     & CV    & CV    & CV \\
    \midrule
    ECT1  & 288   & 0.000     & 0.007 & 0.149 \\
    ECT2  & 864   & 0.000     & 0.001 & 0.002 \\
    ECW08 & 24    & 0.000     & 0.142 & 0.188 \\
    ECW09 & 72    & 0.000     & 0.000     & 0.146 \\
    \bottomrule
    \end{tabular}%
  \label{tab:sen_image}%
\end{table}%

\begin{figure}[htbp]
    \centering
    \includegraphics[width=0.99\columnwidth]{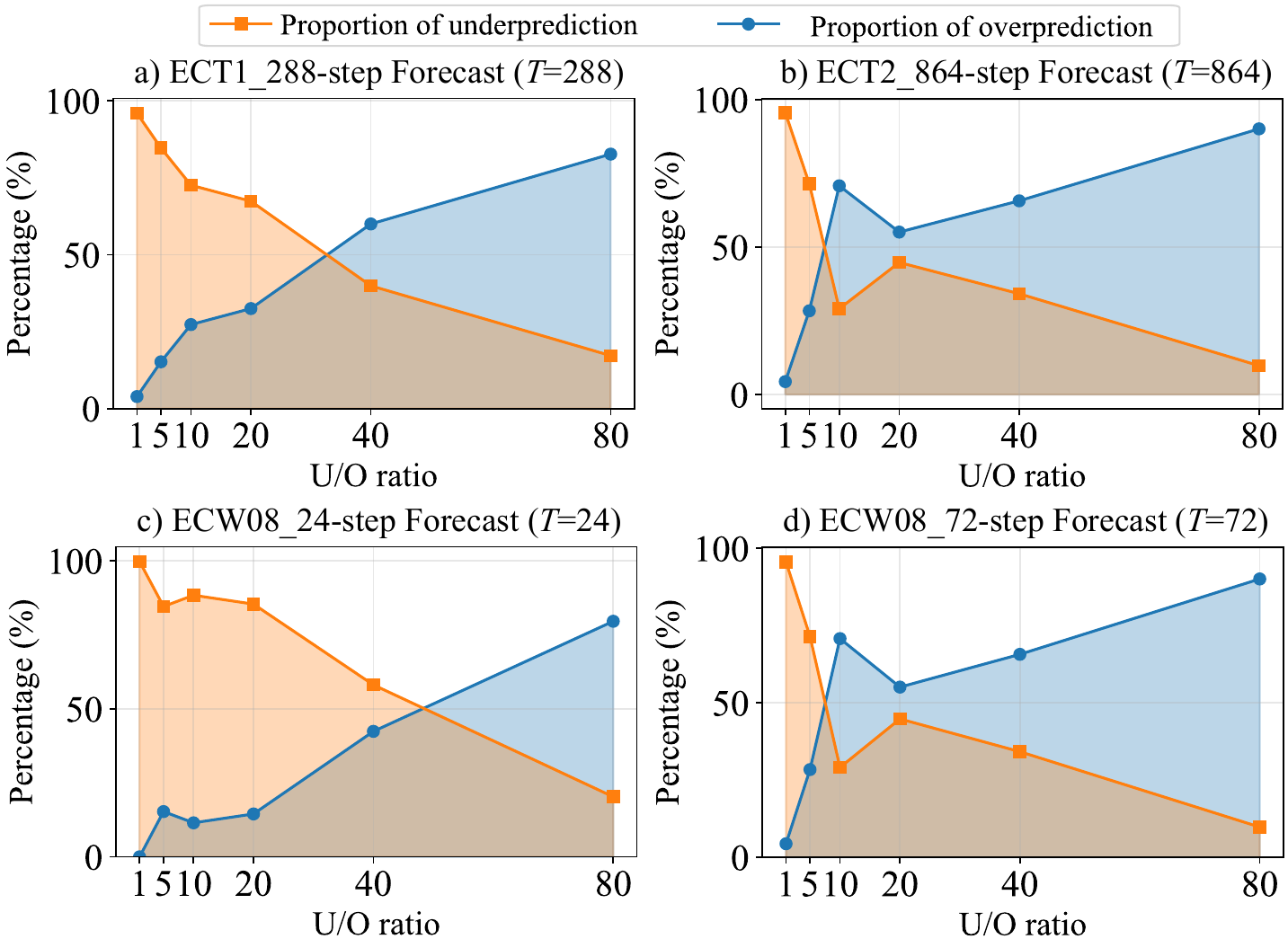}   
    \caption{The proportion of underprediction and overprediction across the entire test set varies with the U/O ratio}
    \label{fig:POL_sen}
\end{figure}

\textbf{Sensitivity of SAL settings. }
To prevent underprediction that may trigger SLA violations, SAL is designed to reflect real-world cost asymmetries. Since the consequences of underprediction and overprediction vary across business scenarios, we define a controllable under-to-over prediction loss ratio (U/O ratio) in SAL, expressed as \(\frac{R+P}{C}\), where \(R+P\) denotes the per unit loss from underprediction, and \(C\) is the loss from overprediction. As shown in Figure~\ref{fig:POL_sen}, the proportion of underprediction and overprediction shifts with the U/O ratio. At U/O = 1, the model treats both errors equally; however, due to frequent traffic spikes, predictions tend to underestimate. As the ratio increases, the model gradually favors overprediction to reduce penalties, with a near balance achieved when the ratio exceeds 20. These results suggest that tuning the U/O ratio enables SAL to adapt to diverse deployment scenarios.

\subsection{Complexity Analysis}
The overall computational complexity of \emph{HRS} scales linearly with the lookback length $L$. This is because all major components, NFEM, FFM, and MDM, consist of linear operations, while VFEM employs fixed-size 2D convolutions on image representations. Owing to constant kernel sizes and strides, the complexity of VFEM also grows linearly with $L$. A detailed derivation is provided in Appendix~\ref{ca}.

\section{Conclusion}\label{con}
This paper presents \emph{HRS}, a hybrid representation framework designed for time series forecasting in Crowdsourced Cloud-Edge Platforms. \emph{HRS} integrates visual and numerical representations to better capture extreme values caused by load surges, and introduces a Scheduling-Aware Loss (SAL) to model the asymmetric impact of prediction errors on downstream decision-making. By aligning forecasting behavior with real-world cost sensitivities, \emph{HRS} enables more effective resource provisioning and supports optimal scheduling decisions. Experiments on four real-world datasets show that \emph{HRS} consistently reduces SLA violations and improves overall profitability.

\begin{ack}
We thank the anonymous reviewers and the area chair for their insightful comments. This work is supported by the National Natural Science Foundation of China (62306208), the Tianjin Natural Science Foundation (23JCQNJC00920), the Key Program of the National Natural Science Foundation of China (U23B2049), the Joint Basic Research Program of Beijing-Tianjin-Hebei Region (F2024201070), and the Tianjin Xinchuang Haihe Lab (22HHXCJC00002).
\end{ack}

%%%%%%%%%%%%%%%%%%%%%%%%%%%%%%%%%%%%%%%%%%%%%%%%%%%%%%%%%%%%%%%%%%%%%%%%

%%% Use this command to include your bibliography file.

\bibliography{HRS}

\newpage
\appendix

\section{Appendix}

\subsection{Representation Transfer Algorithm}\label{algo}

\begin{algorithm}[ht]
\caption{Representation Transfer}
\KwIn{Time series $X \in \mathbb{R}^{L \times M}$, image height $h$, expansion factor $ex$, background color $bc$, line color $lc$, line width $lw$, channel $ch$}
\KwOut{Image array $X_{img}$}

Transpose data: $X \leftarrow X^\top \in \mathbb{R}^{M \times L}$\;

$M \leftarrow$ number of features, $L \leftarrow$ sequence length\;

Initialize $X_{img} \leftarrow \mathbf{0} \in \mathbb{R}^{ch \cdot M \times L \cdot ex \times h \cdot ex}$\;

\For{$i \leftarrow 0$ \KwTo $M-1$}{
    Initialize $img \leftarrow$ RGB image of shape $(h \cdot ex,\; L \cdot ex)$ filled with $bc$\;
    $diff \leftarrow \max(X[i]) - \min(X[i])$\;

    \eIf{$diff = 0$}{
        $line \leftarrow$ vector of ones of length $L$\;
    }{
        $line \leftarrow 1 - \dfrac{X[i] - \min(X[i])}{diff}$\;
    }

    \For{$j \leftarrow 0$ \KwTo $L - 2$}{
        $pt1 \leftarrow (j \cdot ex,\; round(line[j] \cdot (h \cdot ex - 1)))$\;
        $pt2 \leftarrow ((j+1) \cdot ex,\; round(line[j+1] \cdot (h \cdot ex - 1)))$\;
        Draw line from $pt1$ to $pt2$ on $img$ with color $lc$ and width $lw$\;
    }

    Assign $X_{img}[ch \cdot i : ch \cdot (i+1), :, :] \leftarrow img / 255$\;
}

\Return $X_{img}$
\end{algorithm}

\subsection{Measurements from a Commercial CCP} \label{mfac}
\begin{figure}[htbp]
    \centering
    \includegraphics[width=0.99\columnwidth]{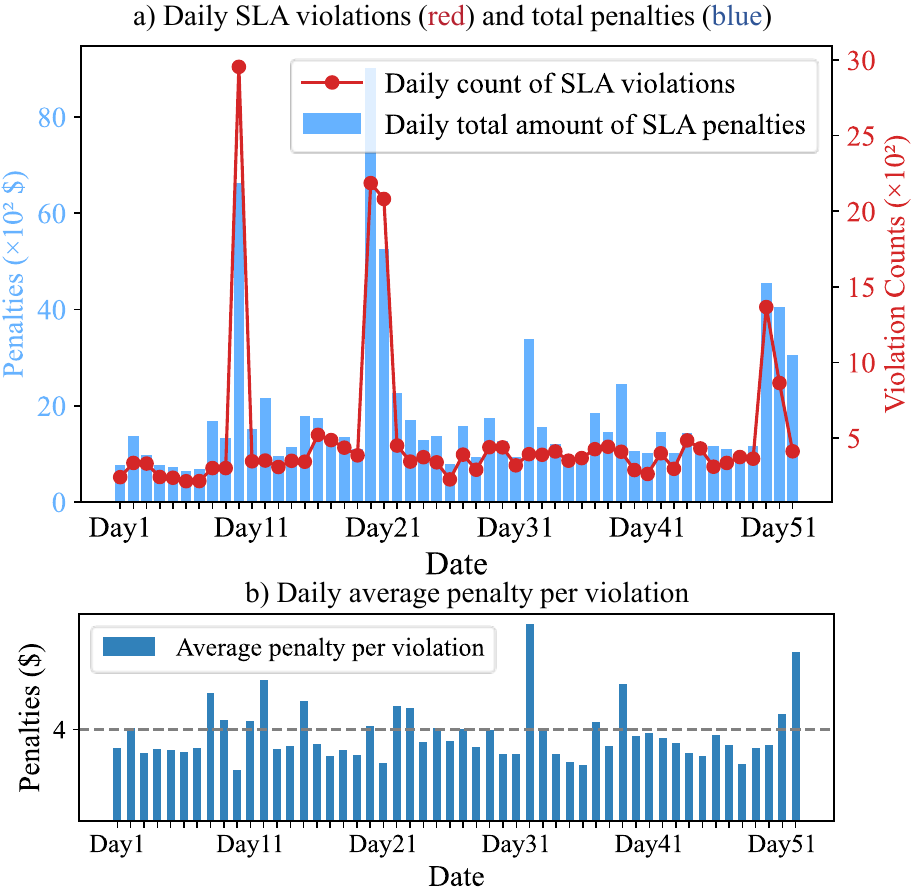}   
    \caption{SLA violation measurements from a CCP.}
    \label{fig:SLA_metr}
\end{figure}

In Crowdsourced Cloud-Edge Platforms (CCPs), the economic impact of underprediction and overprediction is inherently asymmetric. Platforms are obligated to provision sufficient resources and maintain adequate network quality to satisfy task requirements; otherwise, insufficient capacity or performance degradation may result in user churn, compromised QoS, and SLA violations with SMPs, leading to significant financial penalties. As illustrated in Figure~\ref{fig:SLA_metr}, a 52-day trace from one CCP reveals frequent SLA violations, with daily penalties averaging over \$2,000 and individual violations incurring costs of approximately \$4 on average. This underscores the importance of scheduling-aware forecasting to support more effective resource management and minimize operational losses.

\subsection{Use Case} \label{usecase}
To evaluate the practical impact of forecasting quality on scheduling, we simulate a one-month workload using ten servers and real-world datasets. Specifically, ECW datasets are used to emulate server workloads, while ECT datasets represent streaming service demands. A greedy scheduling algorithm is applied based on forecasting results with varying prediction lengths.

\begin{figure}[htbp]
    \centering
    \includegraphics[width=0.95\columnwidth]{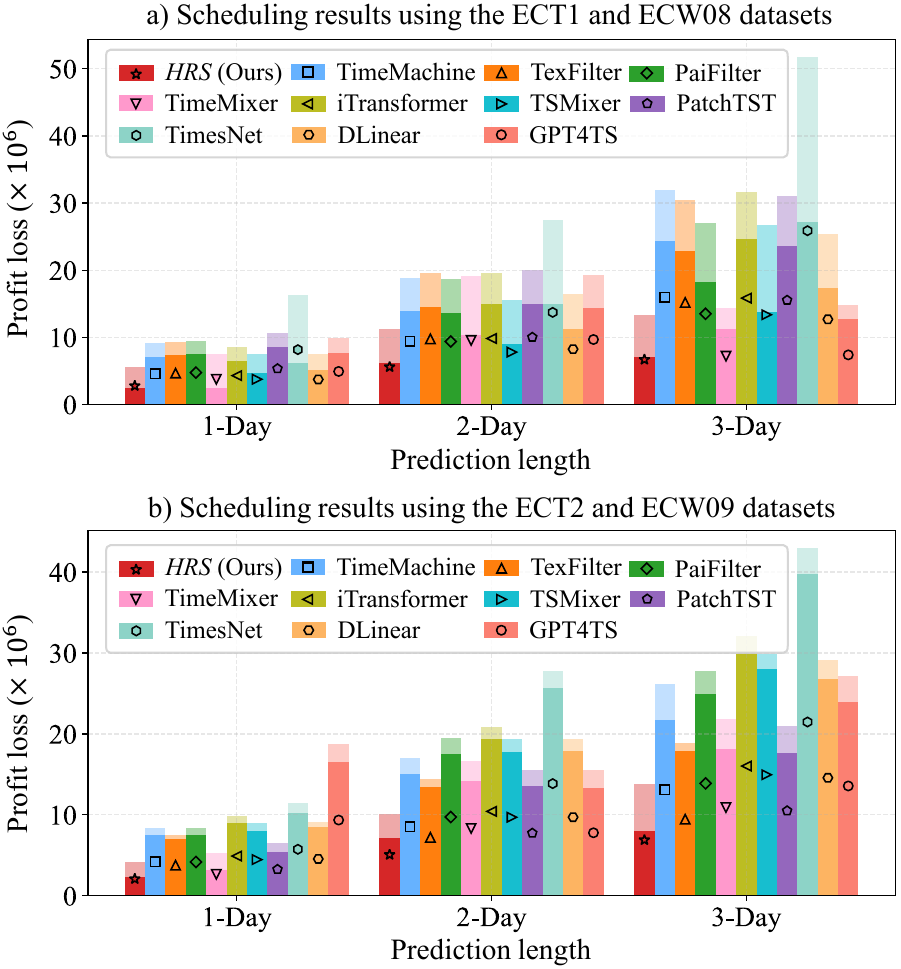}   
    \caption{Profit loss caused by prediction errors in the simulated scheduling results (Darker: loss caused by underprediction, Lighter: loss caused by overprediction).}
    \label{fig:profit_ana}
\end{figure}

Figure~\ref{fig:profit_ana} illustrates the profit loss, capturing the effects of both underprediction and overprediction. Solid bars represent losses due to underprediction, including SLA penalties and revenue loss from unserved traffic. Transparent bars reflect losses from overprediction, corresponding to the cost of redundant resource allocation. \emph{HRS} consistently achieves the lowest total profit loss across all forecasting lengths. Notably, its slightly increased overprovisioning cost effectively avoids the much larger penalties caused by underprovisioning.

\subsection{Visualization}\label{visu}
\begin{figure}[htbp]
    \centering
    \includegraphics[width=\columnwidth]{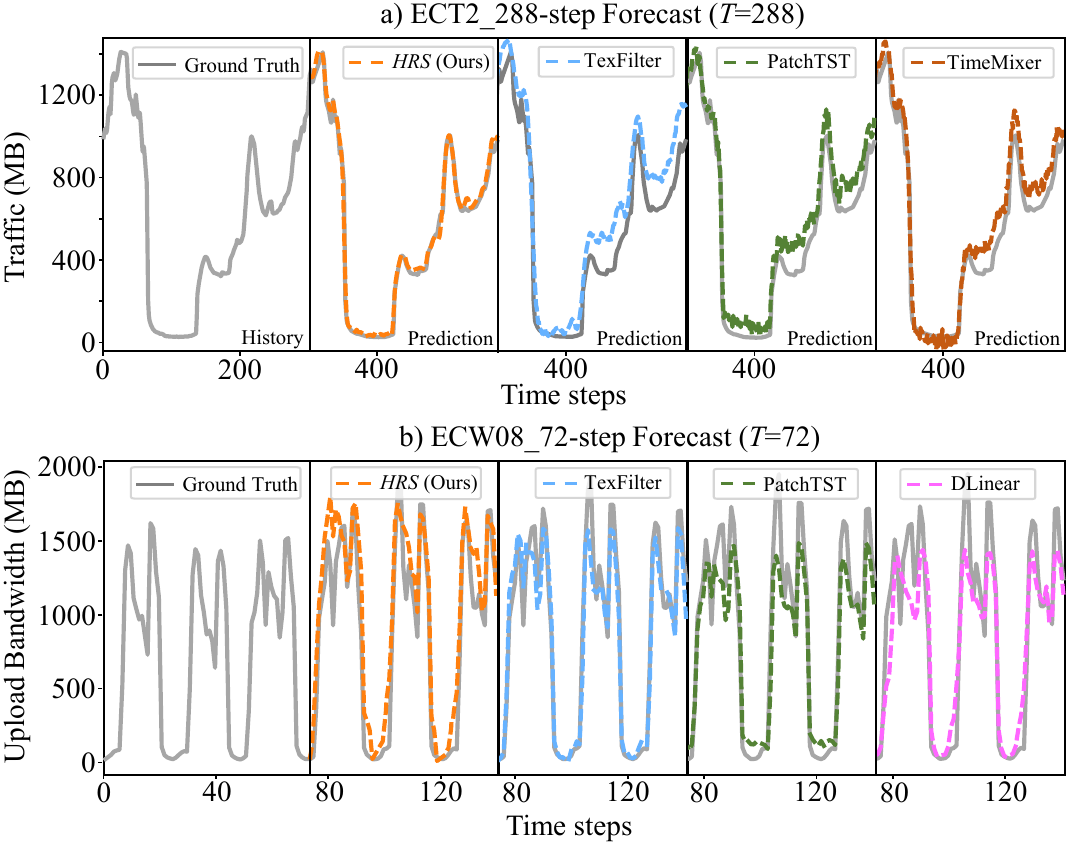}   
    \caption{Prediction results of different samples from ECT2 and ECW08. \emph{HRS} achieves higher accuracy in extreme regions.}
    \label{fig:visual}
\end{figure}

To illustrate how \emph{HRS} benefits from image-based representations in capturing local curve shape, especially in regions where traditional numerical models often fail, such as extreme values, we visualize prediction results on the ECT2 and ECW08 datasets. As shown in Figure~\ref{fig:visual}, we select one sample from each dataset at prediction lengths of 288 and 72, respectively. For comparison, we include the top three baselines (trained with SAL) from Table~\ref{tab:result} and overlay their respective prediction regions. While all models capture the general trend, \emph{HRS} significantly outperforms the baselines in extreme regions, such as peaks and troughs. This superior accuracy ensures that the predicted values provide more reliable guidance for subsequent scheduling decisions.

\subsection{Complexity Analysis} \label{ca}
In latency-sensitive CCP scenarios, algorithms must balance accuracy and efficiency to avoid excessive overhead. This section provides a detailed analysis of complexity of \emph{HRS}.

The complexity of forecasting algorithms is typically closely related to the input sequence length $L$~\cite{pyraformer}. As discussed in Section 4.2, all components of the \emph{HRS} model, except the VFEM, are composed of linear operations (i.e., NFEM, FFM, and MDM). The VFEM processes image inputs of shape $(H, L)$ using a 2D convolution with kernel size $(k_h, k_w)$, stride $(s_h, s_w)$. The resulting feature map has approximate dimensions: $I_h = \left\lfloor \frac{H - k_h}{s_h} \right\rfloor + 1$ and $I_w = \left\lfloor \frac{L - k_w}{s_w} \right\rfloor + 1.$

After reshaping, the extracted shape features $S_f$ grows as $F = O\left( \frac{H \cdot L}{s_h \cdot s_w} \right) $. The linear modules for the time series and images contribute $O(L)$ and $O(F)$ respectively. The overall linear component has a complexity of:
\begin{equation}
    O(L + F) = O\left(L + \frac{H \cdot L}{s_h \cdot s_w}\right) = O\left(L \cdot \left( \frac{H}{s_h \cdot s_w} + 1 \right)\right).
\end{equation}
Since the image height $H$, kernel size, and stride are all fixed constants, the overall computational complexity remains \textbf{linearly dependent} on the input sequence length $L$.

\begin{figure}[htbp]
    \centering
    \includegraphics[width=0.9\columnwidth]{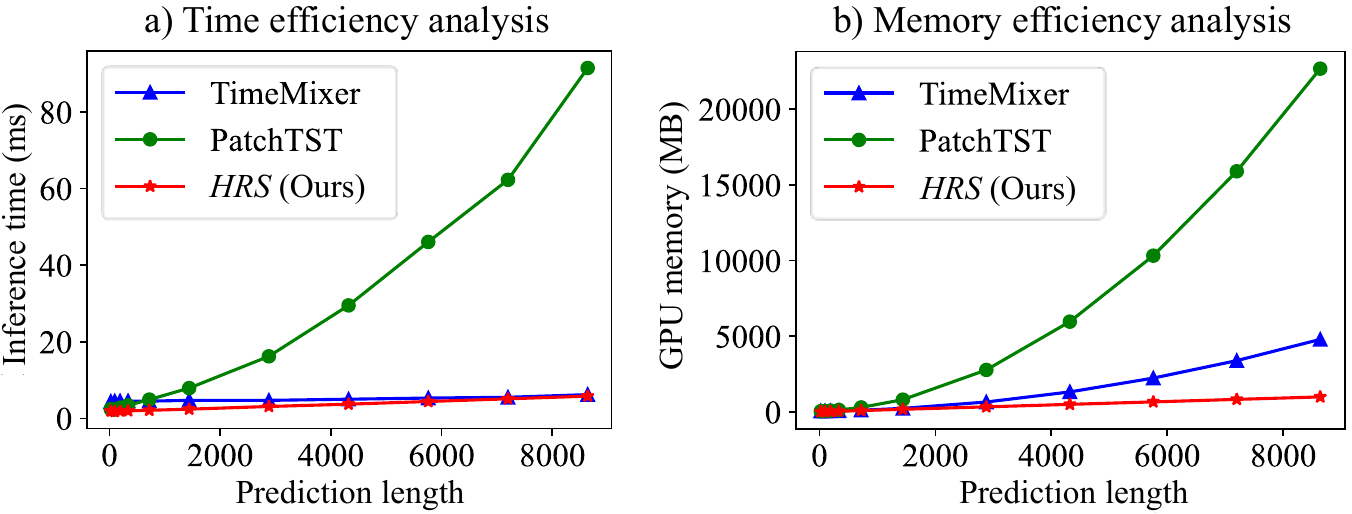}   
    \caption{\emph{HRS} exhibits linear resource and time efficiency}
    \label{fig:time}
\end{figure}

In Figure~\ref{fig:time}, we compare the inference time and GPU memory consumption of \emph{HRS} with the two models that achieved the second-best performance in Table~\ref{tab:result}—TimeMixer and PatchTST, as the prediction length increases. The results show that HRS maintains competitive efficiency compared to the unimodal TimeMixer and PatchTST, primarily due to its core modules being built upon linear architectures. PatchTST, based on a Transformer backbone, suffers from a significant efficiency drop when dealing with very long sequences. In contrast, \emph{HRS} demonstrates better resource efficiency than TimeMixer, despite both adopting linear structures. These characteristics highlight \emph{HRS} as a well-balanced solution that combines accuracy with scalability and computational efficiency, making it a strong candidate for latency-sensitive scenarios such as CCPS.

\subsection{Generalization to Related Domains}\label{gen}
To demonstrate the generalization capability of \emph{HRS} beyond the original scenario, we further evaluate it on the Energy domain using two widely adopted benchmarks, ETTh1 and ETTm1, which present forecasting challenges analogous to those in our paper. We forecast the next 96 time steps and report results in terms of MSE. As shown in Table~\ref{tab:energy_results}, \emph{HRS} consistently achieves the best performance compared to recent state-of-the-art methods. These results suggest that the mechanisms proposed in \emph{HRS} generalize well to similar domains, validating its robustness and transferability.

\begin{table}[h]
\centering
\caption{Results on the Energy domain (lookback length $L = 96$). Baseline results are taken from TimeMachine~\cite{timemachine}.}
\begin{tabular}{lcc}
\toprule
$\mathcal{D}$ → & ETTh1 (5 mins) & ETTm1 (1 hour)\\
${T}$ → & 96 & 96\\
Models & MSE & MSE\\
\midrule
\emph{HRS} (Ours)           & \textbf{0.363}       & \textbf{0.236}       \\
TimeMachine   & 0.364                & 0.317                \\
iTransformer  & 0.386                & 0.334                \\
PatchTST      & 0.414                & 0.329                \\
TimesNet      & 0.384                & 0.338                \\
DLinear       & 0.386                & 0.345                \\
\bottomrule
\end{tabular}
\label{tab:energy_results}
\end{table}

\end{document}